\documentclass{article} 
\usepackage{nips15submit_e,times}
\usepackage{graphicx,epsfig}
\usepackage{subfig}
\usepackage{url, cite,amsmath,bm}

\renewcommand{\vec}[1]{\bm{#1}}

\newcommand{\mat}[1]{\mathbf{#1}}

\newcommand{\vsub}[2]{\vec{#1}_{\text{#2}}}
\newcommand{\msub}[2]{\mat{#1}_{\text{#2}}}
\newcommand{\vsupersub}[3]{\vec{#1}^{(\text{#2})}_{\text{#3}}}
\newcommand{\msupersub}[3]{\mat{#1}^{(\text{#2})}_{\text{#3}}}

\newcommand\Whh{\msub{W}{hh}}
\newcommand\Wxh{\msub{W}{xh}}
\newcommand\WxT{\msub{W}{xT}}
\newcommand\Wxi{\msub{W}{xi}}
\newcommand\Wxf{\msub{W}{xf}}
\newcommand\Wxo{\msub{W}{xo}}
\newcommand\Whi{\msub{W}{hi}}
\newcommand\Whf{\msub{W}{hf}}
\newcommand\Who{\msub{W}{ho}}
\newcommand\Wci{\msub{W}{ci}}
\newcommand\Wcf{\msub{W}{cf}}
\newcommand\Wco{\msub{W}{co}}
\newcommand\Wxc{\msub{W}{xc}}
\newcommand\Whc{\msub{W}{hc}}
\newcommand\Wc{\msub{W}{c}}
\newcommand\mH{\mat{H}}
\newcommand\mT{\mat{T}}
\newcommand\mC{\mat{C}}

\newcommand\ist{\vsub{i}{t}}
\newcommand\xst{\vsub{x}{t}}
\newcommand\yst{\vsub{y}{t}}
\newcommand\fst{\vsub{f}{t}}
\newcommand\cst{\vsub{c}{t}}
\newcommand\ost{\vsub{o}{t}}
\newcommand\hst{\vsub{h}{t}}
\newcommand\hstm{\vsub{h}{t-1}}
\newcommand\cstm{\vsub{c}{t-1}}

\newcommand\hstL{\vsupersub{h}{L}{t}}

\newcommand\cstL{\vsupersub{c}{L}{t}}
\newcommand\cstLp{\vsupersub{c}{L+1}{t}}

\newcommand\cstmLp{\vsupersub{c}{L+1}{t-1}}

\newcommand\cstone{\vsupersub{c}{1}{t}}
\newcommand\cstmone{\vsupersub{c}{1}{t-1}}
\newcommand\xstzero{\vsupersub{x}{0}{t}}
\newcommand\xstone{\vsupersub{x}{1}{t}}
\newcommand\istone{\vsupersub{i}{1}{t}}

\newcommand\dstone{\vsupersub{d}{1}{t}}

\newcommand\dstLp{\vsupersub{d}{L+1}{t}}

\newcommand\xstLp{\vsupersub{x}{L+1}{t}}

\newcommand\bsdLp{\vsupersub{b}{L+1}{d}}
\newcommand\bsdone{\vsupersub{b}{1}{d}}

\newcommand\fstLp{\vsupersub{f}{L+1}{t}}
\newcommand\fstone{\vsupersub{f}{1}{t}}

\newcommand\istLp{\vsupersub{i}{L+1}{t}}
\newcommand\hstmLp{\vsupersub{h}{L+1}{t-1}}
\newcommand\hstmone{\vsupersub{h}{1}{t-1}}

\newcommand\Wxcone{\msupersub{W}{1}{xc}}

\newcommand\Whcone{\msupersub{W}{1}{hc}}

\newcommand\wcdone{\vsupersub{w}{1}{cd}}
\newcommand\Wxdone{\msupersub{W}{1}{xd}}

\newcommand\wldLp{\vsupersub{w}{L+1}{ld}}
\newcommand\wcdLp{\vsupersub{w}{L+1}{cd}}

\newcommand\WxcLp{\msupersub{W}{L+1}{xc}}
\newcommand\WhcLp{\msupersub{W}{L+1}{hc}}
\newcommand\WxdLp{\msupersub{W}{L+1}{xd}}

\title{Depth-Gated LSTM}

\author{
Kaisheng Yao\thanks{Presented at Jelinek Summer Workshop on August 14 2015. }\\
Microsoft Research\\
\texttt{kaisheny@microsoft.com} \\
\And
Trevor Cohn\\
University of Melbourne\\
\texttt{tcohn@unimelb.edu.au} \\
\AND
Katerina Vylomova\\
University of Melbourne\\
\And
Kevin Duh\\
Nara Institute of Science and Technology\\
\And
Chris Dyer\\
Carnegie Mellon University\\
\texttt{cdyer@cs.cmu.edu} 
}

\nipsfinalcopy 

\begin{document}

\maketitle

\begin{abstract}
In this short note, we present an extension of long short-term memory (LSTM) neural networks to using a depth gate to connect memory cells of adjacent layers. Doing so introduces a linear dependence between lower and upper layer recurrent units. Importantly, the linear dependence is gated through a gating function, which we call depth gate. This gate is a function of the lower layer memory cell, the input to and the past memory cell of this layer. We conducted experiments and verified that this new architecture of LSTMs was able to improve machine translation and language modeling performances.
\end{abstract}

\section{Introduction}

Deep neural networks (DNNs) have been successfully applied to many areas, including speech~\cite{Hinton2012} and vision~\cite{Krizhevsky2012}. On natural language processing tasks, recurrent neural networks (RNNs)~\cite{elman1990finding,Jordan1997RNN,Mikolov:2010} are widely used because of their ability to memorize long-term dependency. 

A typical problem of training deep networks, including RNNs, is gradient diminishing and explosion. This problem is apparent when training a simple RNN. 
The long short-term memory (LSTM)~\cite{Hochreiter1997,Gers99learningto} neural networks is an extension of simple RNN~\cite{elman1990finding}. In LSTM, a memory cell has linear dependence of its current activity and its past activity. Importantly, a forget gate is used to modulate the information flow between the past and the current activities. 
LSTMs also have input and output gates to modulate its input and output. 

Perhaps the introduction of gating functions in \cite{Hochreiter1997,Gers99learningto} is the most significant improvement to the recurrent neural networks~\cite{elman1990finding}. 
More recently, the Gated Recurrent Unit~\cite{ChoGRU2014} has also adopted the concept of using gates. LSTMs and GRUs are widely used in many natural language processing tasks~\cite{Bahdanau2014,Sutskever2014}.

To construct a deep neural networks, the standard way is to stack many layers of neural networks. This however has the same problem of building simple recurrent networks. The difference here is that the error signals from the top, instead of from the last time instance, have to be back-propagated through many layers of nonlinear transformations and therefore the error signals might be either diminished or exploded. 

This short note investigates an extension of LSTMs that uses a depth-gate to connect memory cells of the lower and upper layers. We review recurrent neural networks in Sec.~\ref{sec:rnns}. Section~\ref{sec:dglstm} presents the extension. Experiments are in Sec.~\ref{sec:experiments}. We relate this extension with other works in Sec.~\ref{sec:related} and conclude in Sec.~\ref{sec:conclusion}.

\section{Review of recurrent neural networks}
\label{sec:rnns}
A recurrent neural network~\cite{elman1990finding,Jordan1997RNN} has a hidden state $\hst$ that depends on its past value $\hstm$ recursively; i.e.,

\begin{equation}
\hst = g(\Whh \hstm + \Wxh \xst)
\end{equation}
\noindent where $g(\cdot)$ is usually a nonlinear function such as tanh. $\xst$ is the input. $\Whh$ and $\Wxh$ are the weight matrices. 

\subsection{Long short-term memory (LSTM)}
\label{sec:lstm}
\begin{figure}
\centering
  \includegraphics[scale=0.5]{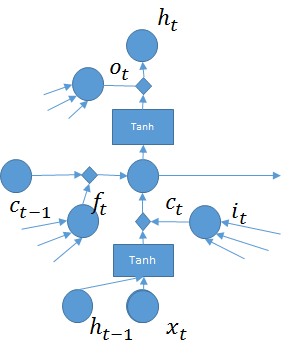}
  \caption{LSTM}
  \label{fig:lstm}
\end{figure}

LSTM was initially proposed in~\cite{Hochreiter1997,Gers99learningto} and later modified in \cite{GravesLSTM:2013}. We follow the implementation in \cite{GravesLSTM:2013}, which is illustrated in Fig.~\ref{fig:lstm}. LSTM  introduces a linear dependence between its memory cells $\cst$ and its past $\cstm$. 
Additionally, LSTM has input and output gates. Specially, LSTM is written below as
\begin{eqnarray}
\ist & = & \sigma(\Wxi \xst + \Whi \hstm + \Wci \cstm) \label{eqn:ig} \\
\fst & = & \sigma(\Wxf \xst + \Whf \hstm + \Wcf  \cstm) \label{eqn:fg} \\
\cst & = & \fst \odot \cstm + \ist \odot tanh(\Wxc \xst + \Whc \hstm) \label{eqn:memory} \\
\ost & = & \sigma(\Wxo \xst + \Who \hstm  + \Wco  \cst ) \label{eqn:og} \\
\hst & = & \ost \odot tanh(\cst) \label{eqn:output}
\end{eqnarray} 
\noindent where $\ist$, $\fst$ and $\ost$ are input gate, forget gate and output gate of LSTM. $\hst$ is the output from the LSTM. $\sigma(\cdot)$ is the logistic function. $\odot$ denotes element-wise product. 
In our application of LSTM, the forget gate and input gate share the same parameters but are computed as $\fst = 1 - \ist$. Note that bias terms are omitted in the above equations but they are applied by default. 

\subsection{Stacked LSTMs}
\label{sec:deeplstm}
Typically, LSTMs are stacked to form deep recurrent neural networks, illustrated in the left figure of Figure~\ref{fig:dglstm}.

The output from the lower layer LSTM at layer $L$ is $\hstL$. With a possible affine transformation, this output is used as input $\xstLp$ in the upper layer LSTM at layer $L+1$. Except for this output-input connection, there is no other connections between the two layers. 

\section{The Depth-gated LSTM}
\label{sec:dglstm}
\begin{figure}[h]
\begin{center}
    \centering
    \subfloat[Stacked LSTM]{{\includegraphics[scale=0.5]{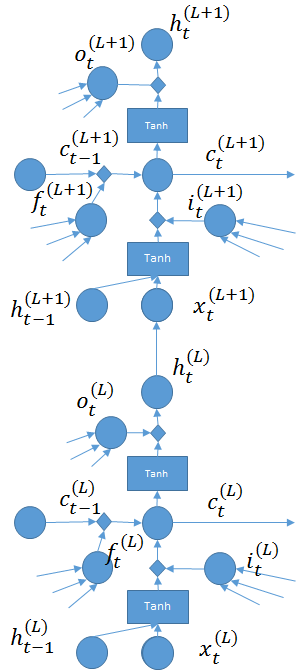} }}%
    \qquad
    \subfloat[Depth-gated LSTM]{{\includegraphics[scale=0.5]{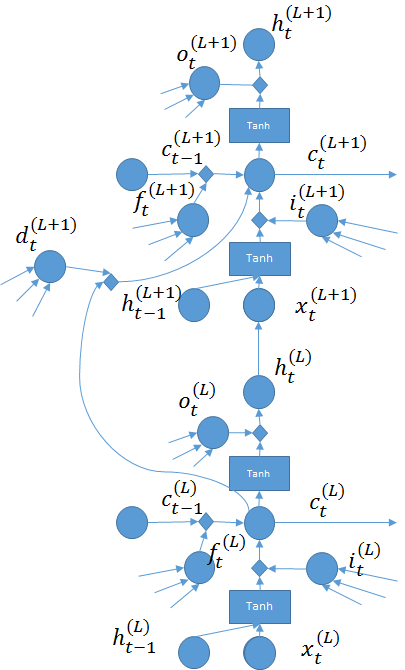} }}%
    \caption{Illustration of the stacked LSTM and the depth-gated LSTM. Notice the additional connection between memory cells in the lower and upper layers in the depth-gated LSTM. }
    \label{fig:dglstm}
\end{center}
\end{figure}

The depth-gated LSTM (DGLSTM)~\footnote{Implementation at https://github.com/kaishengyao/cnn/blob/master/cnn/dglstm.cc and https://github.com/kaishengyao/cnn/blob/master/cnn/dglstm.h.} is illustrated in the right figure of Fig. \ref{fig:dglstm}. It has a depth gate that connects the memory cells $\cstLp$ in the upper layer $L+1$ and the memory cell $\vec{c}_t^L$ in the lower layer $L$. The depth-gate controls how much flow from the lower memory cell directly to the upper layer memory cell. The gate function at layer $L+1$ at time $t$ is a logistic function as 
\begin{equation}
\dstLp = \sigma(\bsdLp + \WxdLp \xstLp + \wcdLp \odot \cstmLp + \wldLp \odot \cstL)
\end{equation}
\noindent where $\bsdLp$ is a bias term. $\WxdLp$ is the weight matrix to relate depth gate to the input of this layer. The past memory cell is also related via a weight vector $\wcdLp$. To relate the lower layer memory, it uses a weight vector $\wldLp$. Note that, if lower and upper layer memory cells have different dimension, $\wldLp$ should be a matrix instead of a vector. 

Using the depth gate, a DGLSTM computes the memory cell at layer $L+1$ as follows
\begin{eqnarray}
\cstLp & = & \dstLp \odot \cstL + \fstLp \odot \cstmLp + \istLp \odot tanh(\WxcLp\xstLp + \WhcLp \hstmLp) 
\end{eqnarray} 

In DGLSTM, equations (\ref{eqn:ig}), (\ref{eqn:fg}), (\ref{eqn:og}) and (\ref{eqn:output}) are the same as the standard LSTM, except that DGLSTM uses a superscript $L+1$ to denote operations at layer $L+1$. 

The idea of using gated linear dependence can also be used to connect the first layer memory cell $\cstone$ with the feature observation $\xstzero$. In this case, the depth-gate is computed for $L=0$ as follows 
\begin{equation}
\dstone  =  \sigma(\bsdone + \Wxdone \xstone + \wcdone \odot \cstmone ),
\end{equation}
and the memory cell is computed as 
\begin{equation}
\cstone = \dstone \odot \left(\Wxdone \xstzero \right) + \fstone \odot \cstmone + \istone \odot tanh(\Wxcone \xstzero + \Whcone \hstmone) 
\end{equation}






\section{Experiments}
\label{sec:experiments}

We applied DGLSTMs on two datasets. The first is BTEC Chinese to English machine translation task. Its training set consists of 44016 sentence pairs. We use its devset1 and devset 2 for validation, which in total have 1006 sentence pairs. We use its devset3 for test, which has 506 sentence pairs. 

The second dataset is PennTreeBank (PTB) for language modeling. It consists of 42075 sentences for training, 3371 sentences for development, and 3762 sentences for test.

\subsection{Machine translation results}

We conducted preliminary experiments and observed that the attention model~\cite{Bahdanau2014} performed better than the encoder-decoder method~\cite{Sutskever2014}. We therefore applied the attention model~\cite{Bahdanau2014} in our experiments. 
Both encoder and decoder used recurrent neural networks in \cite{Bahdanau2014}. However, in this experiment, we only used recurrent neural networks for decoder. For encoder, we used word embedding learned in the training set. 

A preliminary experiment showed that the simple RNN~\cite{elman1990finding} performed the worst. We therefore don't include the simple RNN results in this paper. We compared DGLSTM with GRU and LSTM. All these models used 200-dimension hidden layer. We varied the depth of RNNs. Results in Table~\ref{tab:btecmt} show that DGLSTM outperforms LSTM and GRU in all of the tested depths. 

\begin{table}[t]
\caption{BLEU scores in BTEC Chinese to English machine translation task}
\label{tab:btecmt}
\begin{center}
\begin{tabular}{llll}
\multicolumn{1}{c}{\bf Depth}  & \multicolumn{1}{c}{\bf GRU} & \multicolumn{1}{c}{\bf LSTM} &\multicolumn{1}{c}{\bf DGLSTM}
\\ \hline 
3 & 33.95 & 32.43 & 34.48\\
5 & 32.73 & 33.52 & 33.81 \\
10 & 30.72 & 31.99 & 32.19 
\end{tabular}
\end{center}
\end{table}

In another experiment for the machine translation experiment, we used attention model with DGLSTM to rescore test set k-best lists. We first trained two attention models, one was with 3 layers of DGLSTMs and the other was with 5 layers of DGLSTMs, on training set. Both used 50-dimension hidden layers. We then trained a reranker model using the development data with 100-best lists for each translation pair. The top 100 best lists were generated from the baseline. The features for the reranker models are the scores from the attention model. The 100-best lists on the test set were reranked using the trained reranker model. We ran the above described reranking processes 10 times to get an averaged BLEU scores, which was obtained using one reference. The BLEU scores are listed in Table~\ref{tab:btecmtreranker}. Compared to the baseline, DGLSTM improved BLEU scores by 3 points on the Test set. 
\begin{table}[t]
\caption{BLEU scores by reranking on BTEC Chinese to English  machine translation task}
\label{tab:btecmtreranker}
\begin{center}
\begin{tabular}{lll}
\multicolumn{1}{c}{\bf Dataset}  & \multicolumn{1}{c}{\bf Baseline}  & \multicolumn{1}{c}{\bf DGLSTM}
\\ \hline 
Dev & 26.61 & 30.05\\
Test & 40.63 & 43.08 \\
\end{tabular}
\end{center}
\end{table}

\subsection{Language modeling}
We conducted experiments on PTB dataset. We trained a two layer DGLSTM. Each layer has 200 dimension vector. Test set perplexity results are shown in Table~\ref{tab:lm}. Compared against the previously published results on PTB dataset, DGLSTM obtained the lowest perplexity on PTB test set to our knowledge. 

\begin{table}[t]
\caption{Penn Treebank Test Set Results. }
\label{tab:lm}
\begin{center}
\begin{tabular}{ll}
\multicolumn{1}{c}{\bf Model}  & \multicolumn{1}{c}{\bf Perplexity}
\\ \hline 
RNN~\cite{Mikolov:2010} & 123 \\
LSTM~\cite{GravesLSTM:2013} & 117 \\
sRNN~\cite{Pascanu2014} & 110 \\
DOT(s)-RNN~\cite{Pascanu2014} & 108 \\
DGLSTM & 96
\end{tabular}
\end{center}
\end{table}



\section{Related works}
\label{sec:related}
We developed this method independently in a summer workshop and later knew the works in \cite{HighwayNetworks2015,Kalchbrenner2015GLSTM}. In highway networks in \cite{HighwayNetworks2015}, the output from a layer $\yst$ is a linear function to the input $\xst$, in addition to the output from a nonlinear path. Both of them are gated as follows
\begin{equation}
\yst = \mH(\xst,\Whh) \odot \mT(\xst,\WxT) + \xst \odot \mC(\xst,\Wc)
\end{equation}
\noindent where $\mT$ and $\mC$ are called transform gate and carry gate, respectively.  $\mH(\cdot)$ is the output from a nonlinear path. $\Whh$, $\WxT$, and $\Wc$ are matrices. 
Therefore, the highway network output has a direct and linear connection, albeit gated, to the input. This allows highway networks to train extremely deep networks easily. 

DGLSTM is related to the highway networks in using the same idea of linear and gated connection to input. It differs from the highway networks in not using a specific gate on the non-linear path; DGLSTM keeps the input and output gates which are applied on the non-linear transformations in LSTMs. However, the overall effect of using the input and output gates may be similar to the transfer gate in the highway networks. An additional but important difference is that DGLSTM linearly connects the memory cells in the lower and upper layers. Because of this, the memory cell in DGLSTM has errors back-propagated both from the future and from the top layer, linearly albeit gated. This might be the biggest difference from the highway networks~\cite{HighwayNetworks2015} in its current implementation.


Perhaps the closet work to this research is Grid LSTM~\cite{Kalchbrenner2015GLSTM}, which uses LSTMs in different dimensions and connects them using gated linear connections. Because the dimensions can include not only time, as the typical recurrent neural networks, but also depth and others, Grid LSTM is more general than DGLSTM, which only considers time and depth. Also, Grid LSTM uses a generic form of input, memory, and output. Doing so allows a memory cell to have a gated linear dependence not only on its past memory cell but also on the past observations. Therefore, we consider DGLSTM as a specific and simple case of Grid LSTM that has gate applied to time and depth only on memory cells. However, DGLSTM, Grid LSTM and highway networks share the same idea of stacking networks with both linear but gated connections and nonlinear paths. This idea can be applied to fully connected, convolutional or recurrent layers.

\section{Conclusions}
\label{sec:conclusion}
We have presented a depth-gated LSTM architecture, which uses a depth-gate to have gated linear connection between lower and upper layer memory cells. We observed better performances using this new architecture on machine translation and language modeling tasks. This architecture is related to the highway networks~\cite{HighwayNetworks2015} and Grid LSTM~\cite{Kalchbrenner2015GLSTM} in using an additional linear connection with gates to regulate information flow across layers. 

\section{Acknowledgment}
We thank Juergen Schmidhuber for pointing out highway networks~\cite{HighwayNetworks2015} and grid LSTM~\cite{Kalchbrenner2015GLSTM} works. 

\bibliographystyle{ieeebib}
\bibliography{ref}

\end{document}